\documentclass{article}

\usepackage{PRIMEarxiv}

\usepackage[utf8]{inputenc} 
\usepackage[T1]{fontenc}    
\usepackage{hyperref}       
\usepackage{url}            
\usepackage{booktabs}       
\usepackage{amsfonts}       
\usepackage{nicefrac}       
\usepackage{microtype}      
\usepackage{lipsum}
\usepackage{fancyhdr}       
\usepackage{graphicx}       
\graphicspath{{media/}}     
\usepackage{subcaption}
\usepackage{float}
\usepackage{esvect}
\usepackage{breqn}

\title{Global versus Local: Evaluating AlexNet Architectures for Tropical Cyclone Intensity Estimation 
}

\author{Vikas Dwivedi\\
	Atmospheric Science Research Center\\
	State University of New York, Albany\\
	New York, 12222, USA \\
	\texttt{vdwivedi@albany.edu, vikas.dwivedi90@gmail.com} \\
}

\begin{document}
\maketitle

\begin{abstract}
Given the destructive impacts of tropical cyclones, it is critical to have a reliable system for cyclone intensity detection. Various techniques are available for this purpose, each with differing levels of accuracy. In this paper, we introduce two ensemble-based models based on AlexNet architecture to estimate tropical cyclone intensity using visible satellite images. The first model, trained on the entire dataset, is called the global AlexNet model. The second model is a distributed version of AlexNet in which multiple AlexNets are trained separately on subsets of the training data categorized according to the Saffir-Simpson wind speed scale prescribed by the meterologists. We evaluated the performance of both models against a deep learning benchmark model called \textit{Deepti} using a publicly available cyclone image dataset. Results indicate that both the global model (with a root mean square error (RMSE) of 9.03 knots) and the distributed model (with a RMSE of 9.3 knots) outperform the benchmark model (with a RMSE of 13.62 knots). We provide a thorough discussion of our solution approach, including an explanantion of the AlexNet's performance using gradient class activation maps (grad-CAM). Our proposed solution strategy allows future experimentation with various deep learning models in both single and multi-channel settings. Codes are available at: \url{https://github.com/vikas-dwivedi-2022/Cyclone_Intensity_Estimation_With_AlexNets}
\end{abstract}


\section{Introduction}\label{Sec:Introduction}
A tropical cyclone is a severe low-pressure weather system that forms and intensifies over warm tropical oceans. When a tropical cyclone makes landfall, it can present significant dangers to both human life and property. In the United States, each cyclone routinely results in damages amounting to tens of billions of dollars. The cumulative cost has surpassed 1 trillion dollars since 1980, as indicated by \cite{molina2022social}, and this figure is expected to rise in the future. 

The intensity of a tropical cyclone, measured by the maximum sustained surface wind near its center, is a critical factor in issuing warnings and managing the associated risks and disasters. Table \ref{Tab_Saffir} illustrates the correlation between the destructive power of a tropical cyclone and its maximum sustained wind speed. Please visit \url{https://www.nhc.noaa.gov/aboutsshws.php} for details.

\begin{table}
	\begin{centering}
		\begin{tabular}{|c|c|c|c|c|}
			\hline 
			Level & Name & Type & Wind Speed (knots) & Power\tabularnewline
			\hline 
			\hline 
			1 & Tropical Depression & TD & 0-33 & Weak\tabularnewline
			\hline 
			2 & Tropical Storm & TS & 34-63 & Medium\tabularnewline
			\hline 
			3 & Category 1 & H1 & 64-82 & Medium\tabularnewline
			\hline 
			4 & Category 2 & H2 & 83-95 & Strong\tabularnewline
			\hline 
			5 & Category 3 & H3 & 96-112 & Strong\tabularnewline
			\hline 
			6 & Category 4 & H4 & 113-136 & Strong\tabularnewline
			\hline 
			7 & Category 5 & H5 & >136 & Super Strong\tabularnewline
			\hline 
		\end{tabular}
		\par\end{centering}
	\caption{The Saffir-Simpson wind scale. (1 Knot = 1.85 Km/hr) }\label{Tab_Saffir}
\end{table}

In the existing literature, the problem of detecting tropical cyclone intensity has been addressed through two distinct approaches. The first relies mainly on domain knowledge, while the second takes a purely data-driven approach. 

\begin{figure}[h]
	\centerline{\includegraphics[width=0.7\textwidth]{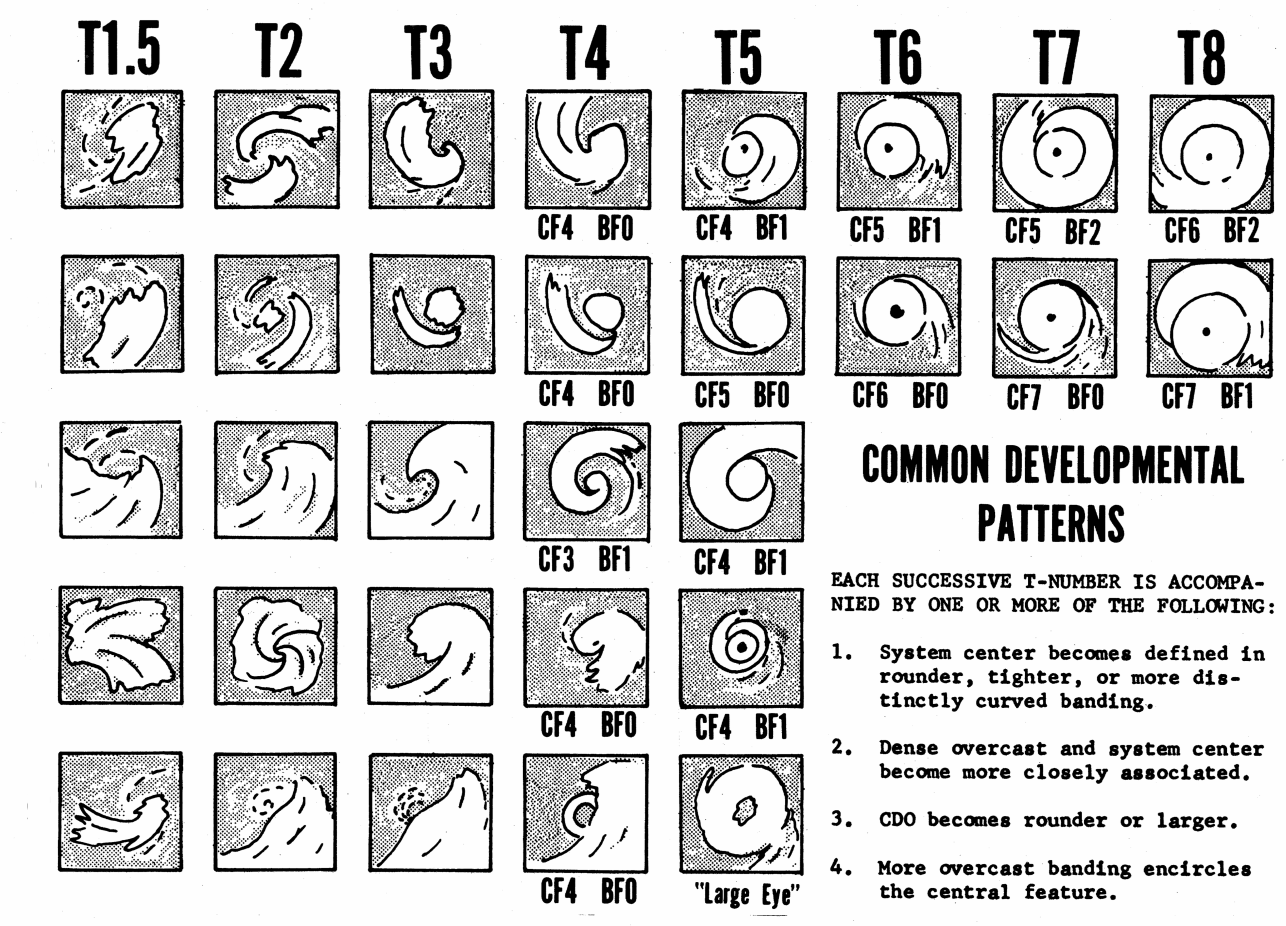}}
	\caption{Common cyclone developmental patterns\cite{dvorak1973technique}.}\label{fig:Dvorak}
\end{figure}

\paragraph{\textbf{Domain Knowledge-Driven Approaches}:} Traditionally, tropical cyclone intensity has been estimated through manual identification of key features in satellite imagery. The classic Dvorak technique \cite{dvorak1973technique,dvorak1975tropical,dvorak1984tropical} exemplifies this approach. Here, meteorologists analyze cloud patterns like central dense overcast and spiral banding, assigning a T-number based on their appearance (Fig. \ref{fig:Dvorak}). These T-numbers correspond to intensity levels like tropical depression, storm, or hurricane. However, this method relies on subjective interpretation and requires prior knowledge of the cyclone's center. To address these limitations, Olander et al. \cite{olander2002development,olander2004advanced} introduced the Advanced Objective Dvorak Technique (AODT), which automates cyclone center detection and wind speed calculation. A further advancement, the Advanced Dvorak Technology (ADT) \cite{olander2007adt,olander2019advanced}, has become the dominant method for real-time intensity estimation. Another category of techniques, termed DAV-T, relies on variance of deviation angle. Pineros et al. \cite{pineros2008objective,pineros2010detecting} introduced the Deviation Angle Variation Technique (DAV-T), based on the premise that stronger tropical cyclones exhibit better cloud symmetry. This technique highlighted the significant finding that the variance in deviation angles correlates with tropical cyclone intensity. Additionally, there are techniques \cite{kar2019tropical,wang2021conformal} based on geometrical and textural features of input images. These features include mean, variance, density, area of cyclone, area of the eye of the cyclone, etc. Finally, ensemble-based techniques like SATCON \cite{velden2019update,velden2020consensus} combine intensity estimates from various infrared and microwave-based methods \cite{herndon20044d} to produce a more robust consensus estimate. Finally there are ensemble-based techniques like SATCON \cite{velden2019update,velden2020consensus} which objectively combines intensity estimates from multiple infrared and microwave-based \cite{herndon20044d} techniques to produce a consensus tropical cyclone intensity estimate.
\paragraph{\textbf{Data-Driven Approaches}:} In contrast to domain-driven methods, the feature extraction process in data-driven method is entirely automated, with a convolutional neural network (CNN) serving as the primary tool for this purpose. Various CNN-based architectures employed for this task can be broadly classified into two groups: (A) Regression-based CNNs and (B) Classification-based CNNs. The primary distinction between the two lies in the choice of loss function. Classification-based CNNs utilize the cross-entropy loss function for multi-class classification, while regression-based CNNs employ loss functions such as mean square error. Two main algorithms for regression-based CNNs are \textit{Deep-PHURIE} \cite{dawood2020deep} and \textit{Deepti} \cite{maskey2020deepti}. \textit{Deep-PHURIE} is trained on infra-red images and \textit{Deepti} is trained on visible images. The authors of \textit{Deepti} also provided a physical explanation of working of CNN by highlighting important portions of images using Grad-CAM \cite{selvaraju2017grad}. Therefore, we will also be comparing our results against \textit{Deepti}. Chen at al. \cite{chen2018rotation} have showed that performance of regression-based CNN can be further improved by combining satellite image channels information like visible light, infrared, water vapor and passive microwave. For classification-based CNNs, readers are encouraged to study the earliest paper by Pradhan et al. \cite{pradhan2017tropical} and a more recent one by Wang et al. \cite{wang2021tropical}. There are several other papers, such as by Chen et al.\cite{chen2020novel} and Yu et al.\cite{yu2020novel}, in which cyclone intensity estimation is considered as two successive tasks, classification for intensity categorization and regression for wind speed estimation. 

\paragraph{\textbf{Hybrid Approaches}:}There are some recent papers, for example by Xu et al. \cite{XU2023105673} and Higa et al. \cite{Higa2021}, which combine domain knowledge with data-driven methods. In both of these papers, the main idea is automatic zooming into the central region of the cyclone in the input images.

\paragraph{\textbf{Contributions:}} We have introduced two ensemble-based models utilizing the AlexNet architecture to estimate tropical cyclone intensity from visible satellite images. The global model is trained across all cyclone categories of the Saffir-Simpson scale, while the local models are trained on specific categories. Both these models outperform the deep learning benchmark model \textit{Deepti}\cite{maskey2020deepti} on a publicly available dataset. In short, the main highlights of this paper are as follows:
\begin{enumerate}
	\item \textbf{Novelty}: The distributed AlexNet architecture employs a two-step regression approach, distinguishing it from existing data-driven models that directly utilize regression or classification followed by regression. This concept draws inspiration from the well-known Mixture of Experts (MoE) models \cite{Masoudnia2014}.  
	\item \textbf{Explainability}: The publicly available cyclone dataset is part of Driven Data competition for cyclone wind speed prediction. The benchmark RMSE for the competition was 13.62 knots and the RMSE of the top three models are 6.26, 6.42, 6.46 knots respectively (\url{https://github.com/drivendataorg/wind-dependent-variables}).  A prevalent strategy among these solutions was the utilization of large ensembles comprising advanced deep learning models such as VGG, ResNet, Inception, etc. While it's true that ensembling hundreds of models may enhance accuracy, such complex methods are not explainable. We have explained our models by showing a direct connection between its gradient class activation maps (grad-CAM) and the cloud structures (Fig.\ref{fig:Dvorak}) discussed in Dvorak’s celebrated empirical model. 
	\item \textbf{Application}: We have clearly explained our solution methodology addressing issues such as data imbalance, batch sample imbalance, temporal correlation, and overfitting which would be helpful for the readers. They can easily tweek the model by introducing multiple-channel inputs for exploiting temporal information or additional information like moisture, brightness temperature, etc.  
\end{enumerate}

\paragraph{\textbf{Organization:}}
The structure of this paper is outlined as follows: Section \ref{Sec:Methodology} gives the details of the methodology. Section \ref{Sec:Results and Discussion} contains the discussion of the results, and Section \ref{Sec:Conclusion} concludes the paper.

\section{Methods}\label{Sec:Methodology}
\subsection{Dataset}\label{subsec:Dataset}
The training dataset comprises single-band satellite images from 494 distinct storms in the Atlantic and East Pacific Oceans, each accompanied by its respective wind speeds. These images are taken at different points in the life cycle of each storm. The dataset can be downloaded from Driven Data competition website (\url{https://www.drivendata.org/competitions/72/predict-wind-speeds/page/327/#dataset-link}). 

Each image measures 366 by 366 pixels. The training dataset consists of 70,257 images paired with corresponding wind speed labels, while the test dataset consists of 44,377 images. Additionally, the dataset contains temporal and categorical features represented by timestamps and ocean identifiers. However, similar to the reference deep learning model \textit{Deepti}, we also do not use the temporal and categorical data.
\begin{figure}[h]
	\centerline{\includegraphics[width=0.8\textwidth]{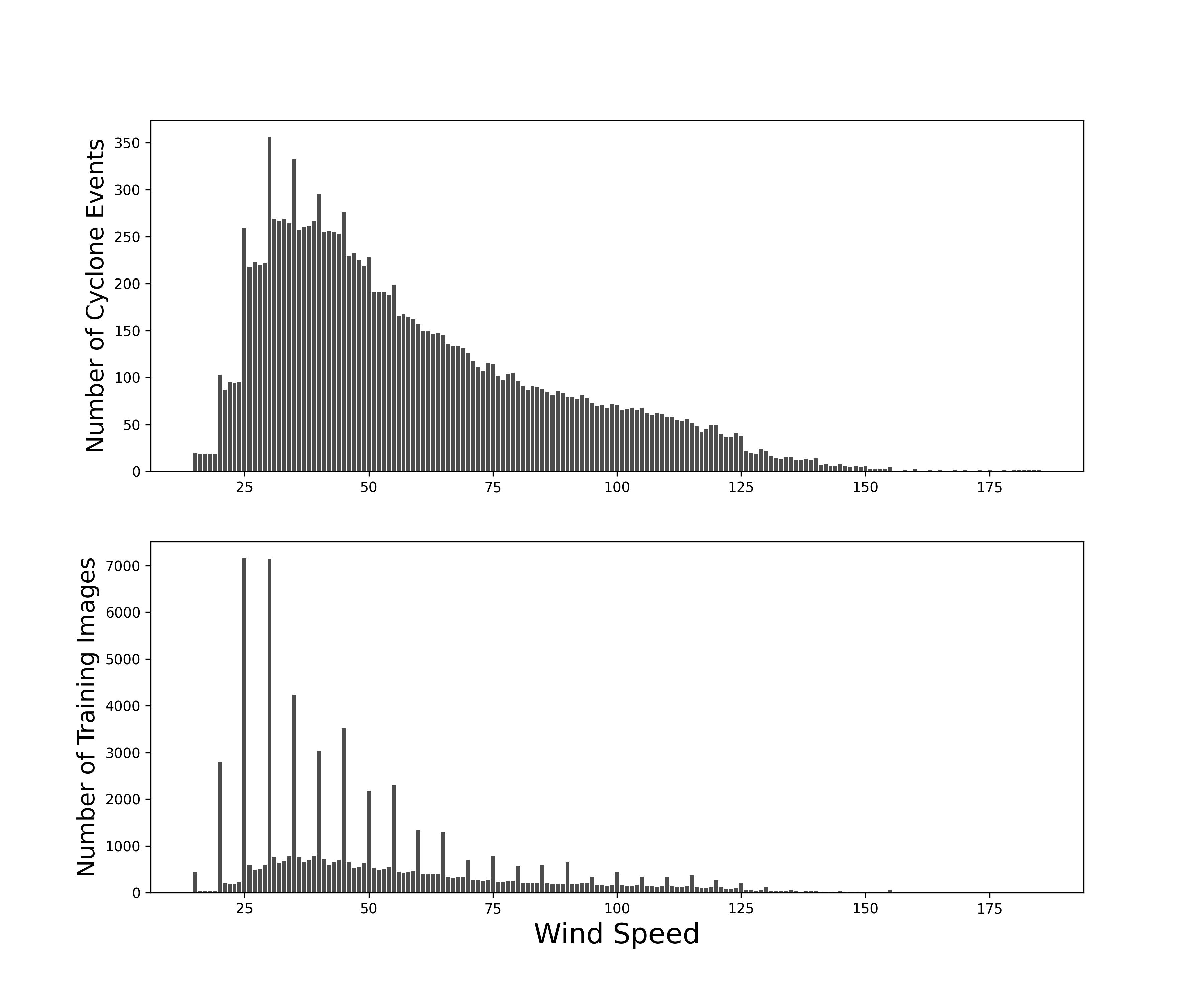}}
	\caption{Event bias (top) and wind speed bias (bottom) in the training dataset.}\label{fig_data_bias}
\end{figure}

\subsubsection*{Dataset Imbalance} The training dataset is highly biased in terms of wind speeds and cyclone events due to the following: 
\begin{itemize}
	\item The total number of storms is 494. However, the number of images per unique storm varies between 4 to 648.
	\item The wind speeds range from 15 to 185 knots but a majority of images fall between 30 and 62 knots. 
\end{itemize}

In Fig. \ref{fig_data_bias}, the heavy-tailed distributions arising from these biases are depicted. The upper sub-figure illustrates the non-uniformity in the count of unique cyclone events per unique wind speed, while the lower sub-figure reveals the uneven distribution of data points, i.e., images per unique wind speed. 

\subsubsection*{Tackling Data Imbalance} To tackle imbalance issues, we adopt the following sampling strategy:
\begin{enumerate}
	\item \textbf{Remove temporal correlation:} Given that certain cyclone events are associated with a significantly large number of images, random splitting between the training and validation sets may lead to a substantial temporal correlation between the two. To address this, we adopt a data-splitting strategy where we ensure that there are no common events shared between the training and validation sets. This approach aims to mitigate potential temporal correlations, enhancing the independence of the two sets during model training and evaluation.  
	\item \textbf{Remove wind speed bias:} The wind speeds span from 15 to 185 knots, with a predominant concentration between 30 and 62 knots. To ensure a comprehensive representation of all wind speeds in our training batch, we included one image corresponding to each unique speed. This approach guarantees that our training data encompasses the full spectrum of wind speed values, leading to a more balanced and inclusive model.
	\item \textbf{Remove event bias:} The number of images per storm exhibit a substantial range for a given wind speed. When randomly selecting an image for a specific wind speed, there is a high probability that the chosen image predominantly originates from an event with a large number of images. To mitigate this bias, our approach involves two steps: first, we randomly select the event for a given speed, and subsequently, we randomly chose an image from the selected event. This method ensures that our batches remain unbiased toward events with a higher count of data points. 
	\item \textbf{Perform data augmentation}: To improve generalization, almost 50 percent of the training batch is augmented with image rotations as well.
\end{enumerate}

\subsection{Models}\label{subsec:Models}
\subsubsection{Global AlexNet Model}
\begin{figure}[h]
	\centerline{\includegraphics[width=0.99\textwidth]{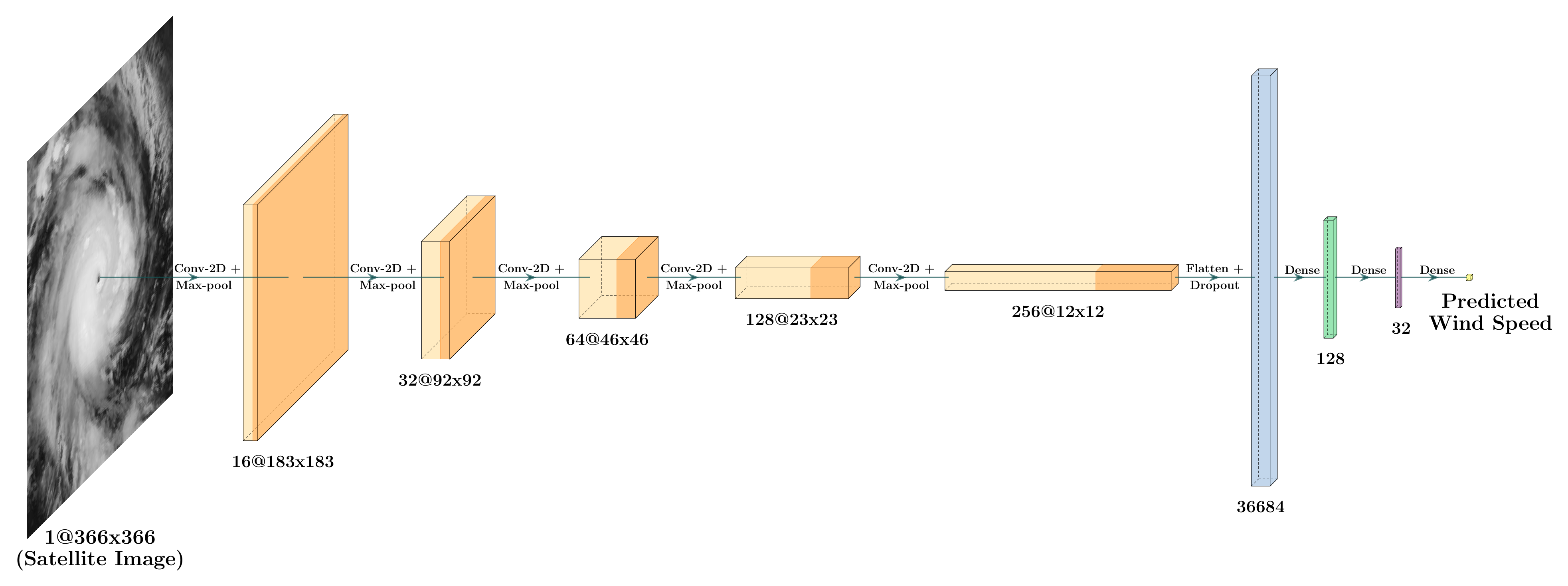}}
	\caption{Architecture of the global Alexnet model. Plot generated using PlotNeuralNet(\url{https://github.com/HarisIqbal88/PlotNeuralNet})}\label{fig_alexnet}
\end{figure}

The network architecture, inspired by the renowned AlexNet model, is illustrated in Fig. \ref{fig_alexnet}. Comprising seven layers, the first five are convolutional layers, each succeeded by max-pooling and batch normalization layers, while the last two are fully connected layers. In all layers, the filter shapes for convolutional and max-pool filters are (3,3) and (2,2), respectively. The Rectified Linear Unit (ReLU) is the activation function. Dropout and regularization are incorporated to prevent overfitting, resulting in a total of almost 5 million training parameters. 

The loss function is Mean Square Log Error (MSLE) and the optimizer used is Adam. The rationale for using the difference between the logarithms of predicted ($\hat{y}$) and true values (\textit{y}), as opposed to the difference in the values themselves, lies in the robustness of MSLE towards outliers. MSLE is defined by $$\text{MSLE} = \frac{1}{N} \sum_{i=1}^{N} (log(\hat{y}_i) - log(y_i))^2,$$\textit{N} being the number of training examples. 
\subsubsection{Distributed AlexNet Model}
\begin{figure}[h]
	\centerline{\includegraphics[width=0.8\textwidth]{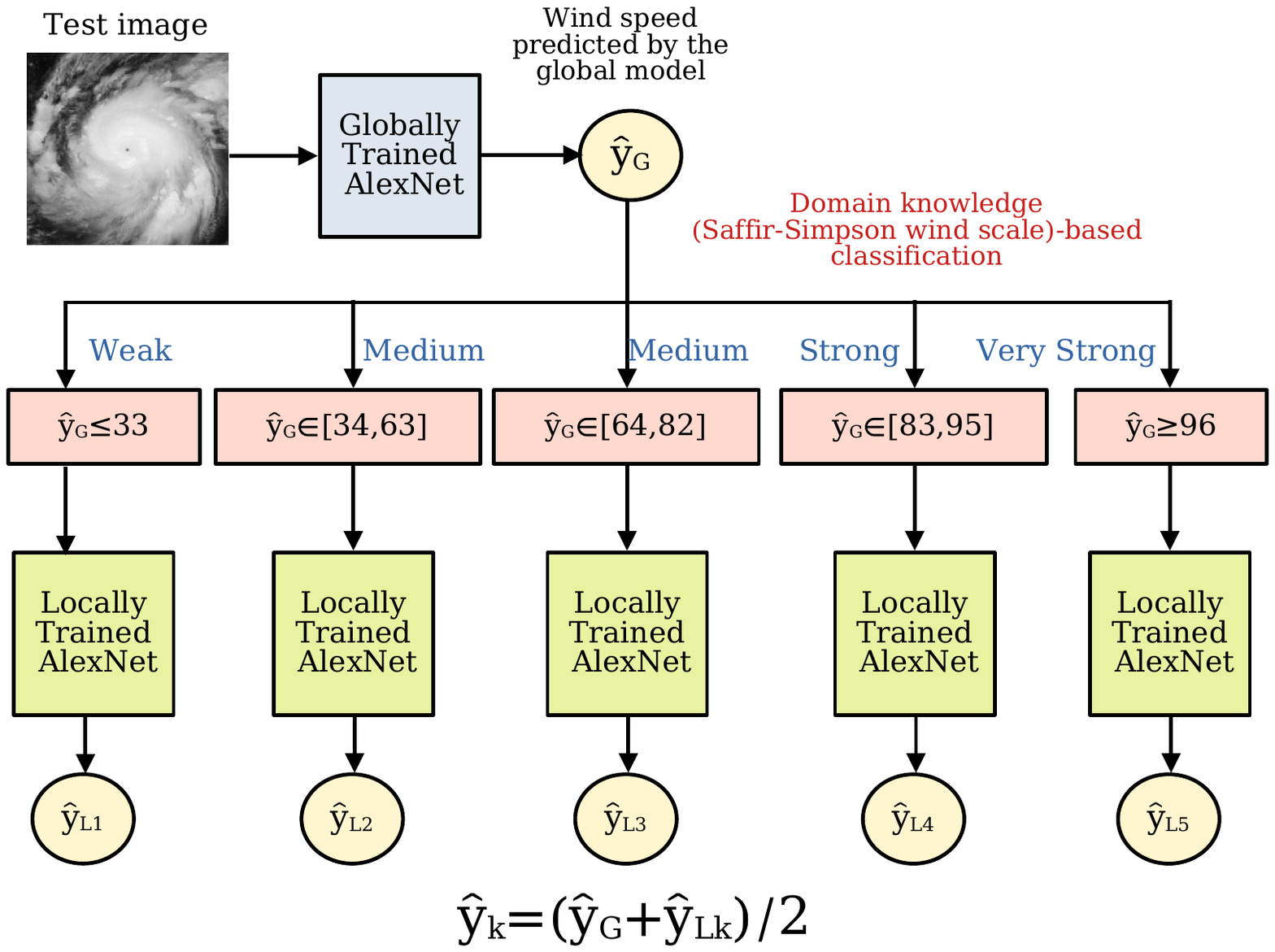}}
	\caption{Schematic of the distributed Alexnets. The final output is the average of global and local models.}\label{fig_dist_alexnet}
\end{figure}

The schematic diagram of the distributed AlexNet model is shown in Fig. \ref{fig_dist_alexnet}. It consists of two layers of neural networks. The initial layer represents the global AlexNet itself. The subsequent layer comprises multiple local AlexNets, each trained on distinct categories delineated in the Saffir-Simpson scale. In the first phase, a satellite image is inputted into the global AlexNet, which produces the wind speed as an output. Following this, based on the wind speed obtained, we identify the category to which the satellite image belongs. Subsequently, the satellite image is once again fed into a local AlexNet, specifically trained on satellite images of the identified category. The resulting wind speed from the distributed AlexNet is the average of the wind speeds predicted by the global and local AlexNets. The distributed architecture draws inspiration from the Mixture of Experts (MoE) model \cite{Masoudnia2014}. In MoE terms, the global AlexNet can be likened to the router or gating function, while the local AlexNets can be seen as individual experts.
\section{Results and Discussion}\label{Sec:Results and Discussion}
In this section, we evaluate the effectiveness of both global and distributed models in estimating cyclone intensity.
Next, we utilize grad-CAM-generated patterns to show (1) the significance of ensemble techniques and (2) their correlation with traditional Dvorak's patterns. All experiments are executed in a Tensorflow 2.13.1 environment, operating on a Lambda vector workstation equipped with three NVIDIA RTX A6000 GPUs, AMD Threadripper processors, and 1 TB of memory.

\paragraph{\textbf{Evaluation Metric:}} Given a training dataset comprising single-band satellite images (\textit{X}) paired with their respective wind speeds (\textit{y}), if the global/local model generates wind speed predictions ($\hat{y}$), the evaluation metric is the Root Mean Square Error (RMSE), which is defined by the formula: $$\text{RMSE} = \sqrt{\frac{1}{N} \sum_{i=1}^{N} (\hat{y}_i - y_i)^2},$$\textit{N} being the number of images in the test dataset. The performance benchmark established by the Driven Data competition is the \textit{Deepti model}, which achieves an RMSE of 13.62 knots (\url{https://www.drivendata.org/competitions/72/predict-wind-speeds/}). Hence, all comparisons are conducted relative to this benchmark. Some other metrics used in the literature are as follows:
\begin{enumerate}
	\item Mean absolute error (MAE): $\frac{1}{N}\sum_{i}|\hat{y}_{i}-y_{i}|$
	\item Bias: $\frac{1}{N}\sum_{i}(\hat{y}-y_{i})$
	\item Relative RMSE: $\frac{1}{avg(\hat{y})}\sqrt{\frac{\sum(\hat{y}_{i}-y_{i})^{2}}{N-1}}$
\end{enumerate}

\subsection{Performance Evaluation}
\begin{itemize}
	\item {\textbf{Ensemble of Global AlexNets}}: We utilize multiple iterations of the global AlexNet model, each trained on different subsets of the complete training data, chosen randomly with replacement. Afterwards, we average the predictions of these models. Specifically, we trained 10 instances of the AlexNet model. The root mean square error (RMSE) for the training and validation datasets is close to 4.65 knots. It is important to emphasize that all subsets of the training data encompassed full range of wind speeds. 
	
	The performance of the global AlexNet model is illustrated in Fig. \ref{pred_ens}. RMSE for the test dataset is recorded at 9.03 knots, surpassing the performance of the benchmark \textit{Deepti} model. Additionally, this global model functions as the gating mechanism for the distributed model. When categorizing the original and predicted wind speeds using the Saffir-Simpson scale, the resulting confusion matrix is presented in Fig. \ref{pred_router}. The key performance parameters—precision, recall, and F1 score—are 0.75, 0.73, and 0.74 respectively.
	\begin{figure}[h]
		\centering
		\begin{subfigure}{0.48\textwidth}
			\centering
			\includegraphics[width=\textwidth]{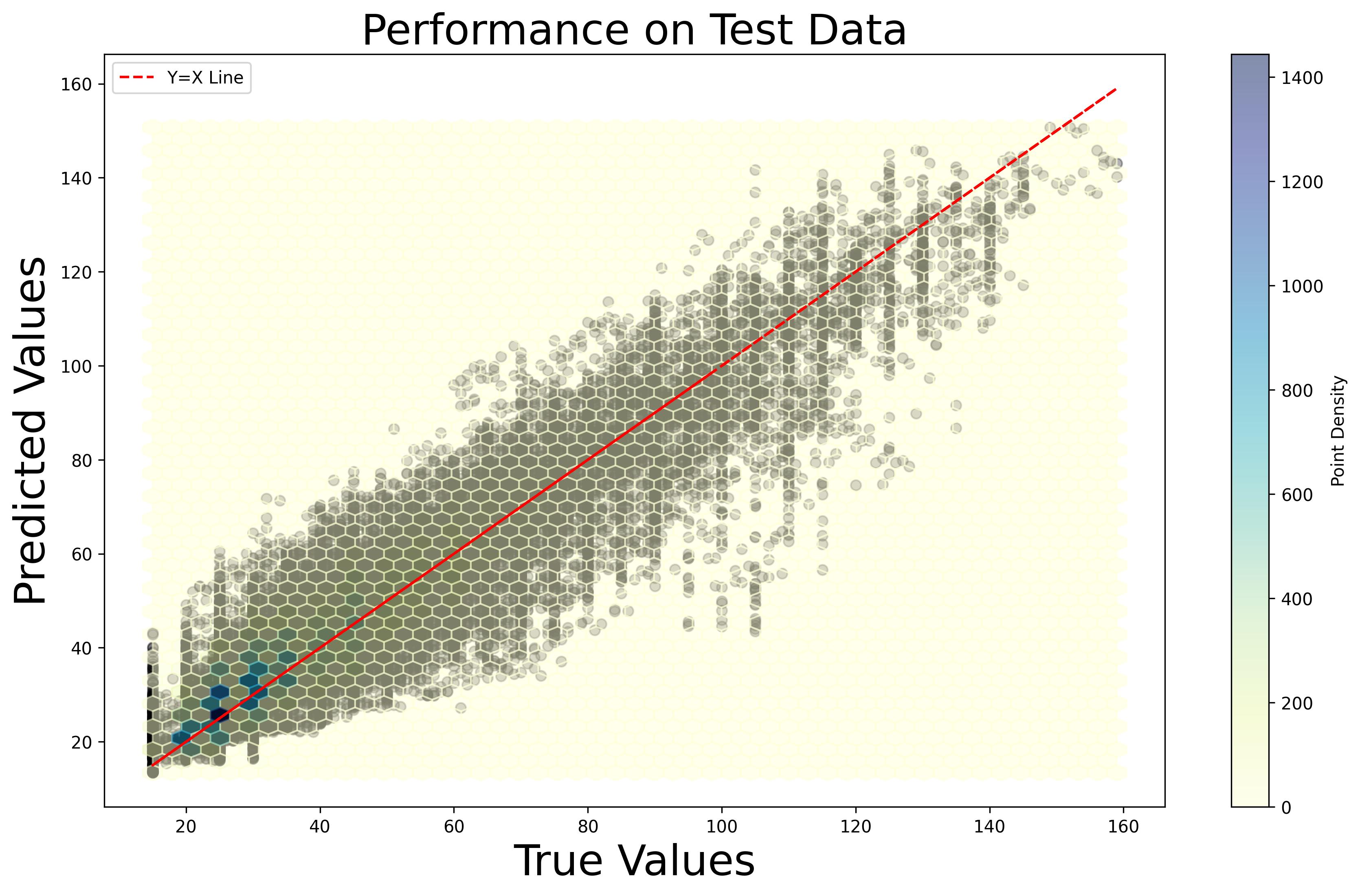}
			\caption{Predictions made by the global AlexNet model across the complete test dataset. RMSE: 9.03 knots.}
			\label{pred_ens}
		\end{subfigure}
		\hfill
		\begin{subfigure}{0.48\textwidth}
			\centering
			\includegraphics[width=\textwidth]{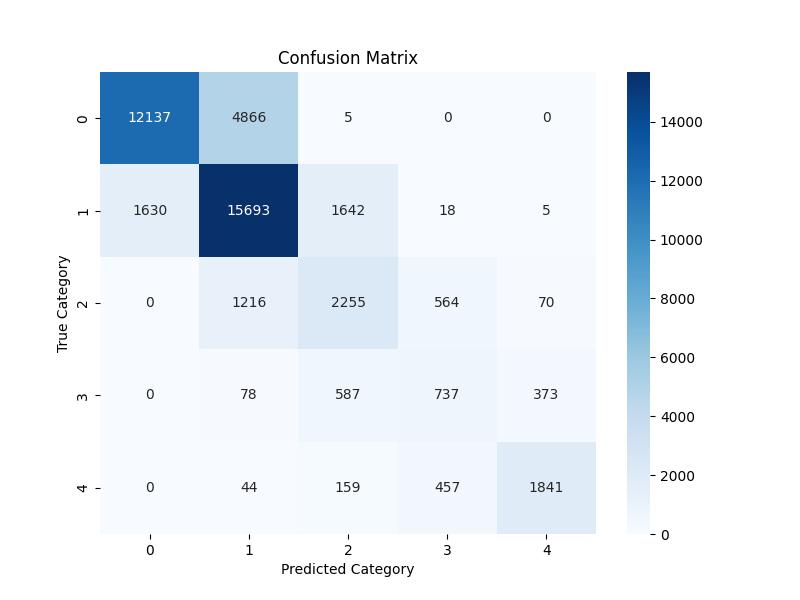}
			\caption{Performance of the gating function on the test dataset. Precision, recall, and F1 score: 0.75, 0.73, and 0.74, respectively.}
			\label{pred_router}
		\end{subfigure}
		\caption{Performance of the global model in regression and classification.}
	\end{figure}
	\item {\textbf{Distributed AlexNets}}: Distributed AlexNets involve training individual models or experts on distinct categories of the Saffir-Simpson wind scale. Initially, we trained these models on entirely non-overlapping subsets. However, this approach resulted in overfitting problems at the boundaries. As a result, we subsequently trained the models on intersecting datasets, where the intersection comprised one-third of the wind speed range of adjacent Saffir-Simpson category. Figures \ref{fig:pred_both} and \ref{fig:pred_both_ovfix} show the performance of distributed AlexNets trained on non-intersecting and intersecting subsets respectively. The root mean square errors for the two cases are 9.5 and 9.3 knots respectively. The full comparison between  the global and local models is given in Table \ref{Tab_Compare}.
	
	\begin{figure}[h]
		\centering
		\begin{subfigure}{0.48\textwidth}
			\centering
			\includegraphics[width=\textwidth]{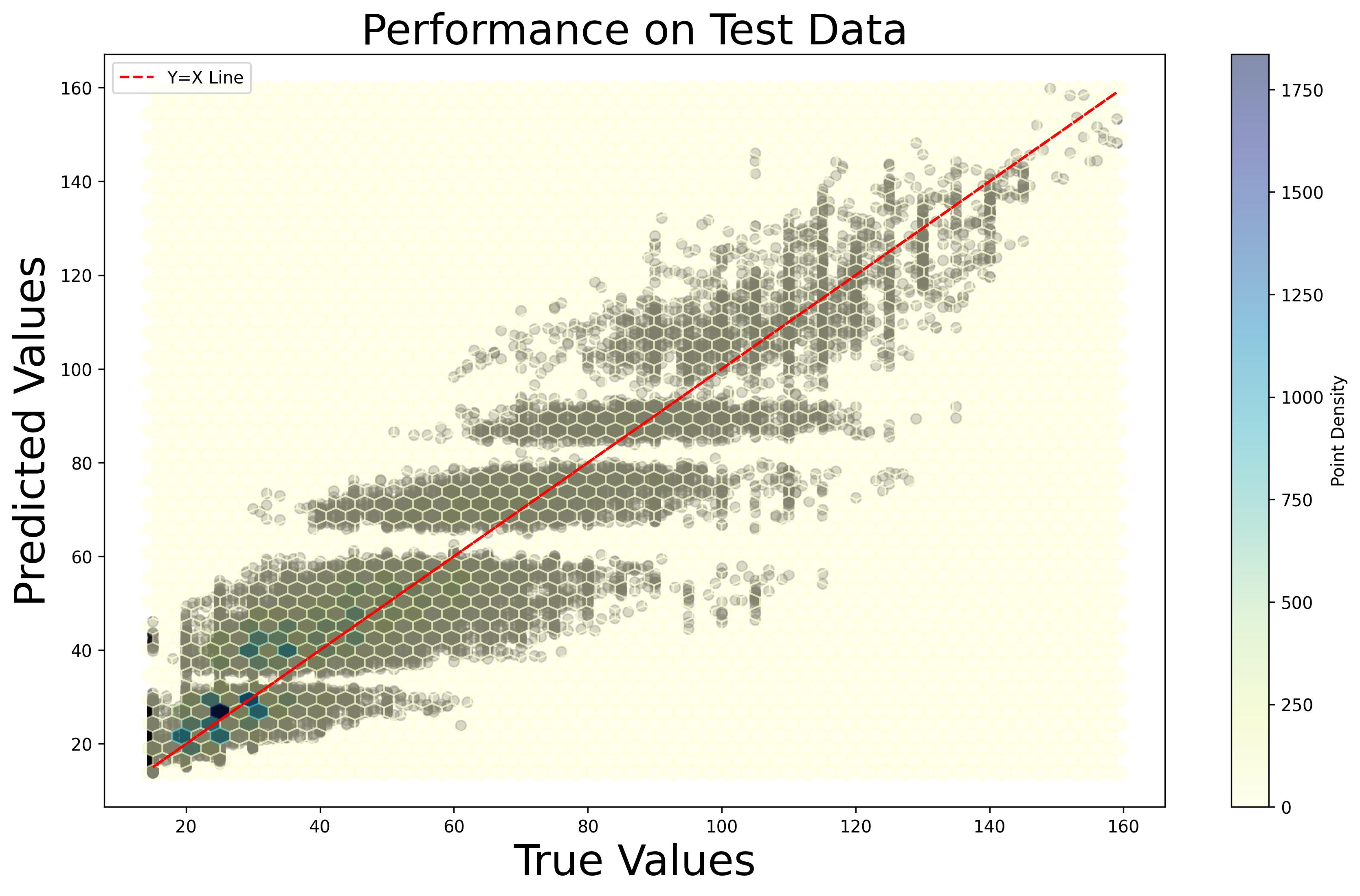}
			\caption{Without overlapping boundaries. RMSE: 9.5 knots.}
			\label{fig:pred_both}
		\end{subfigure}
		\begin{subfigure}{0.48\textwidth}
			\centering
			\includegraphics[width=\textwidth]{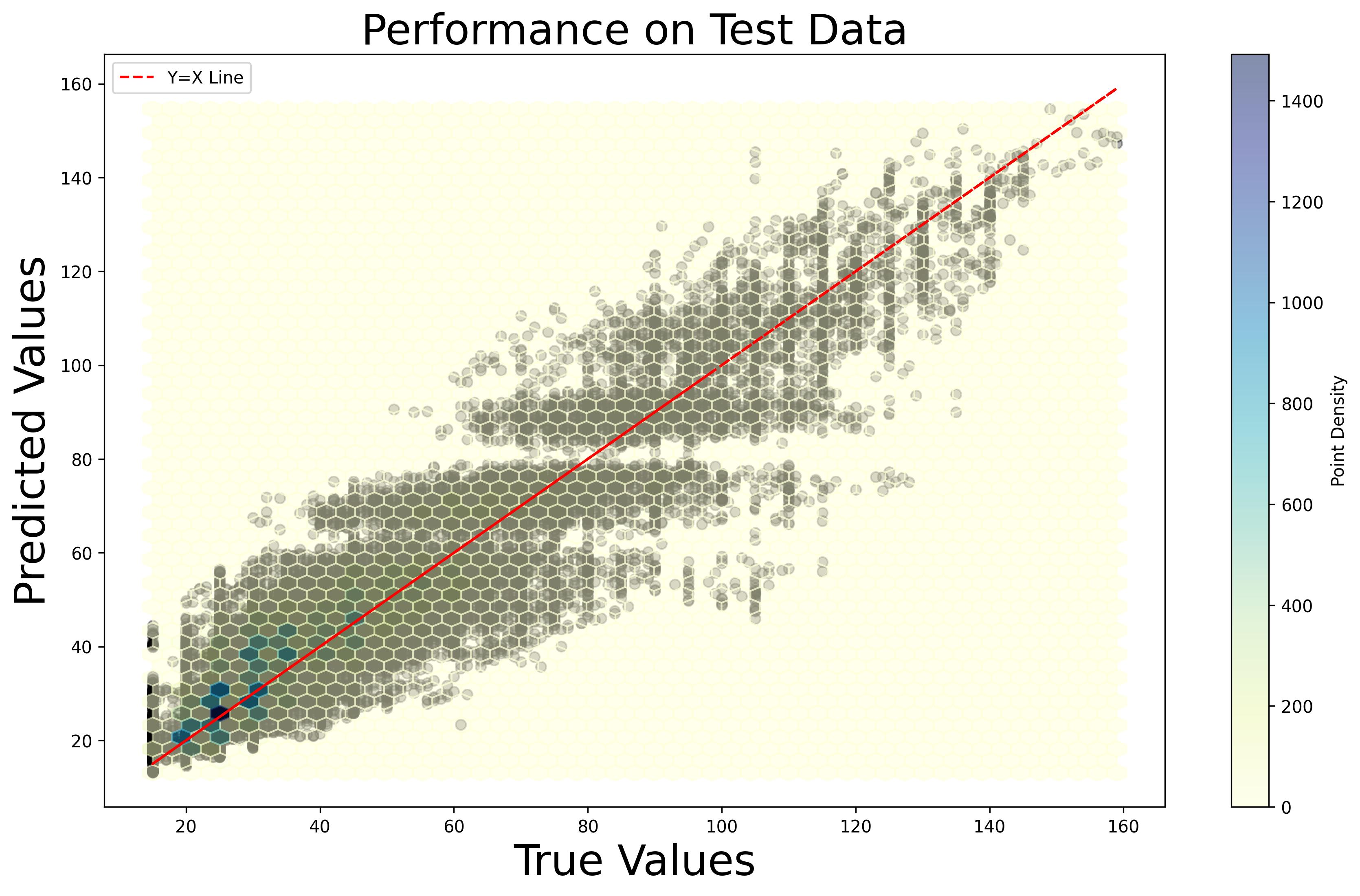}
			\caption{With overlapping boundaries. RMSE: 9.3 knots.}
			\label{fig:pred_both_ovfix}
		\end{subfigure}
		\caption{Predictions made by the distributed AlexNet model across the complete test dataset.}
	\end{figure}
\end{itemize}
\begin{table}
	\begin{centering}
		\begin{tabular}{|c|c|c|}
			\hline 
			Evaluation Metric & Global AlexNet & Local AlexNets\tabularnewline
			\hline 
			\hline 
			RMSE & 9.03 & 9.38\tabularnewline
			\hline 
			MAE & 6.67 & 6.97\tabularnewline
			\hline 
			Bias & 1.20 & 1.52\tabularnewline
			\hline 
			Relative RMSE & 0.17 & 0.17\tabularnewline
			\hline 
		\end{tabular}
		\par\end{centering}
	\caption{Comparison between global and local models on test dataset.}\label{Tab_Compare}
	
\end{table}
\subsection{Model Explanation using Grad-CAM}
As deep learning models grow in complexity, understanding and tracing how the algorithm arrived at a conclusion becomes increasingly challenging for humans. The entire computation process transforms into a "black box," making interpretation impossible. Consequently, it is important to have a set of procedures and techniques that enable human users to understand and trust the results and outputs generated by machine learning algorithms. 

In our case, we explain the functionality of AlexNet by comparing grad-CAM patterns with the cloud pattern, as proposed in the renowned Dvorak's model. A similar methodology has previously been employed by Maskey et al. \cite{maskey2020deepti} for explaining Deepti algorithm. 

\begin{itemize}
	\item Figure \ref{filter_mask} depicts an instance of a grad-CAM pattern overlaid on the original cyclone image. The left sub-figure exhibits the original satellite image. In the middle sub-figure, the median of the masks generated by the component models is displayed. The individual heat maps are presented in Fig. \ref{heatmaps}. Finally, the right sub-figure demonstrates grad-CAM capturing the characteristic comma shape as outlined by Dvorak.
	\begin{figure}[h]
		\centerline{\includegraphics[width=0.75\textwidth]{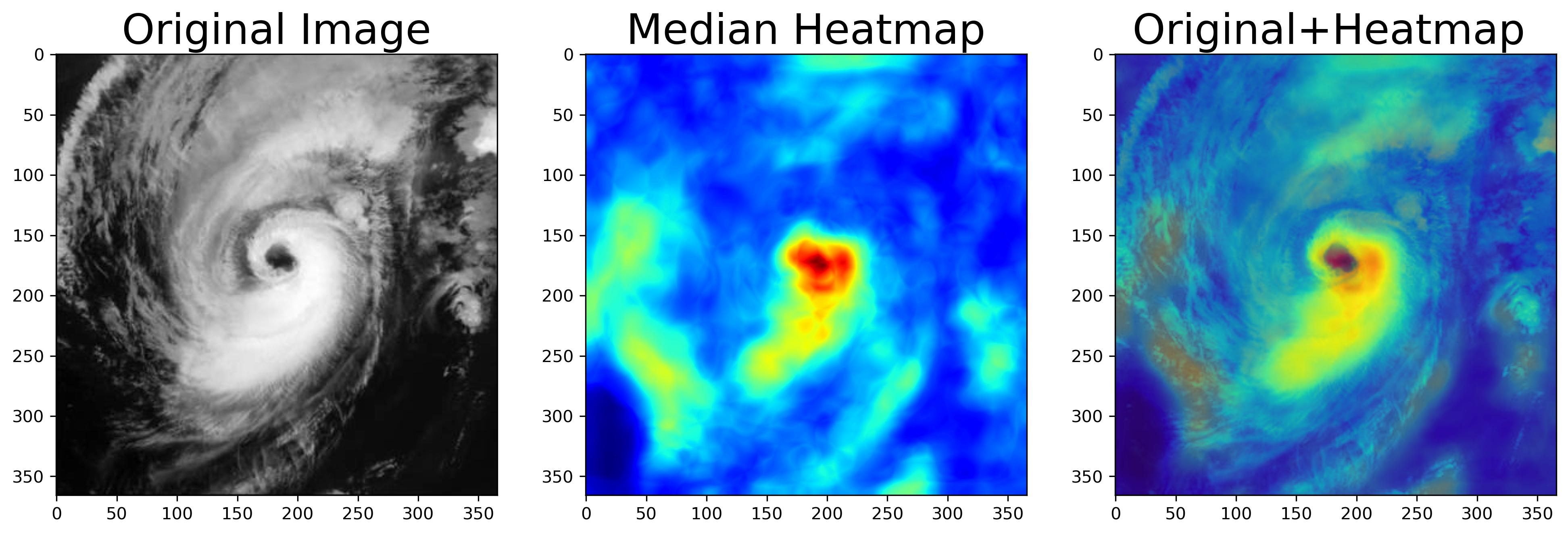}}
		\caption{An exmaple of grad-CAM pattern for a cyclone image.}
		\label{filter_mask}
	\end{figure}
	\item Figure \ref{heatmaps} shows the individual heat maps. Please note that the diversity observed in the individual heat maps indicates that the bootstrapped models exhibit variability and emphasize distinct regions when generating predictions. This is a desirable property because if the models are all identical, there is no point in combining them.
	\begin{figure}[h]
		\centerline{\includegraphics[width=0.65\textwidth]{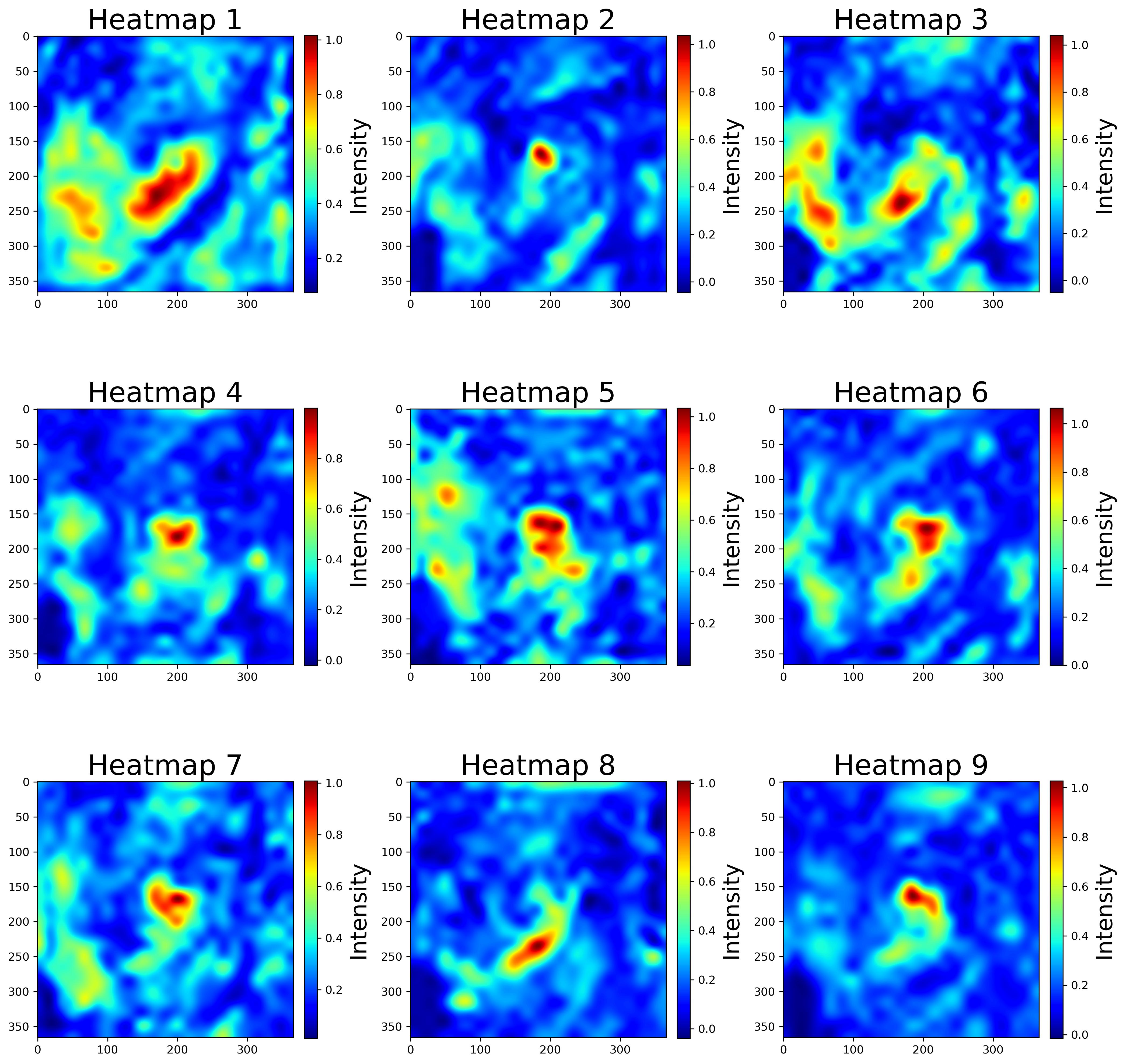}}
		\caption{Heatmaps generated by the global Alexnet models of the ensemble.}\label{heatmaps}
	\end{figure}    
	\item For high-speed cyclones, meteorologists emphasize the well-developed eye as the most crucial cloud feature for estimating wind speed. Figure \ref{pred_eye} displays various images of high-speed cyclones alongside the corresponding heat maps generated by grad-CAM. These heat maps adeptly focus on the cyclone's eye in high-speed scenarios. Overall, the features produced by grad-CAM closely correspond with domain knowledge, indicating that the cloud structure near the cyclone center gains greater importance as wind speed rises.
	\begin{figure}[h]
		\centerline{\includegraphics[width=0.6\textwidth]{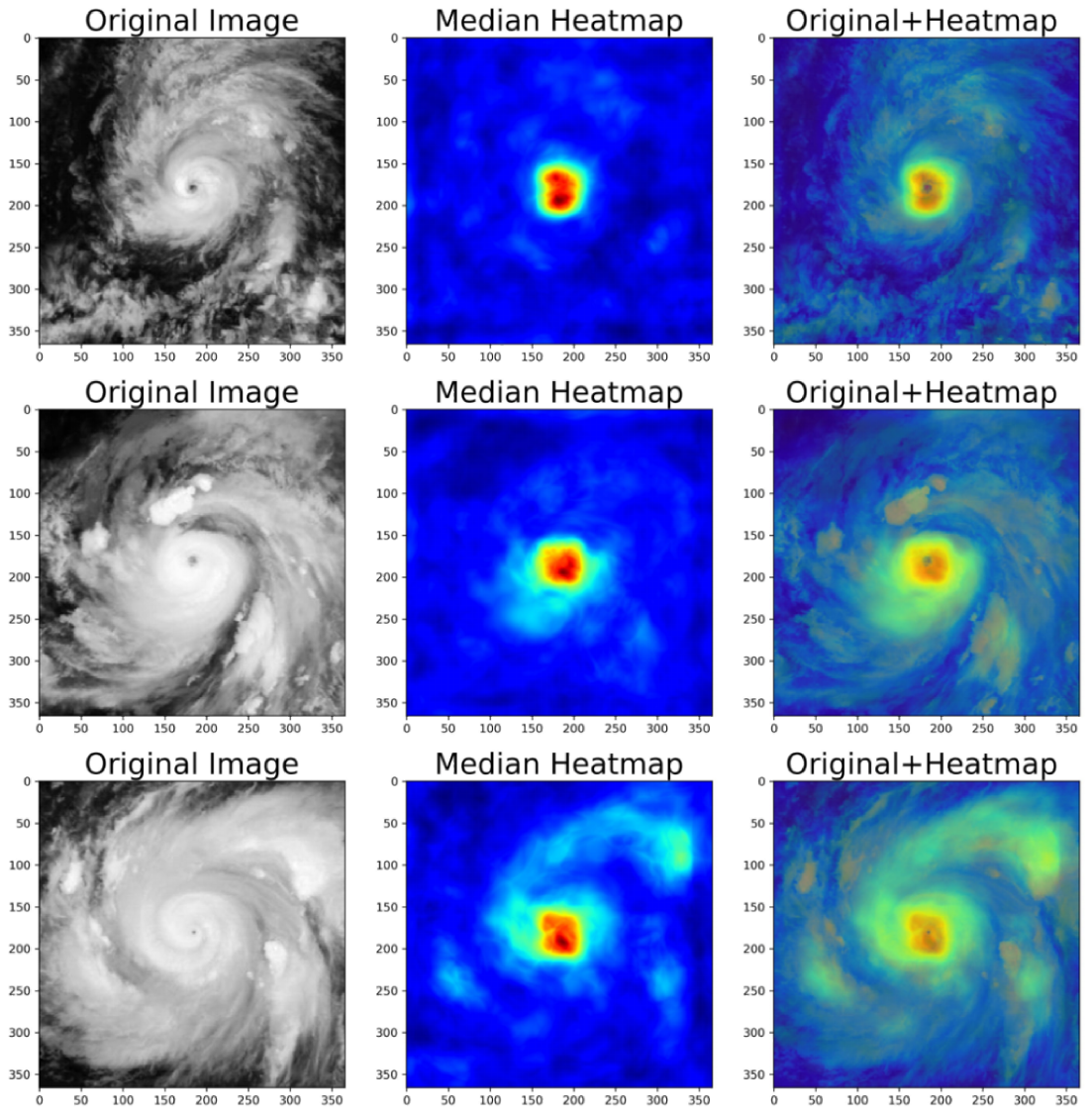}}
		\caption{Grad-CAM intensity of AlexNet is maximum near the "eye" region of high speed cyclones.}\label{pred_eye}
	\end{figure}
\end{itemize}

\paragraph{\textbf{Future Work:}}
In this research, we have assessed solely the performance of AlexNet, primarily due to its minimal memory and computing requirements. However, there is potential for exploring other deeper models such as VGG \cite{simonyan2015deep}. VGG may be a more suitable option, particularly as a gating mechanism, given its superior classification performance compared to AlexNet \cite{Verdhan2021}. Importantly, our methodology does not limit us from incorporating temporal information. Therefore, we can also incorporate multi-channel inputs in future works. 
\section{Conclusion}\label{Sec:Conclusion}
In this paper, we have evaluated two ensemble-based models leveraging the AlexNet architecture. The first model, termed the global AlexNet model, was trained on the entire dataset, while the second model adopted a distributed or local approach in which multiple instances of AlexNet were trained separately on subsets of the training data categorized according to the Saffir-Simpson wind speed scale. Both the models outperformed the deep learning benchmark model \textit{Deepti} on a publicly available cyclone image dataset.  We have clearly discussed the training data imbalance and overfitting issues, and suggested remedies for the same. We have also explained our models by showing a direct connection between its gradient class activation maps (grad-CAM) and the cloud structures discussed in Dvorak’s celebrated empirical model. In conclusion, the ensemble-based models introduced in this paper demonstrate promising potential for improving cyclone intensity estimation using visible satellite images. Further research in this direction, including use of deeper networks and incorporation of multi-channel inputs, can improve their accuracy and reliability.

\section*{Acknowledgements}
The author expresses sincere gratitude to Professor Sukanta Basu of the Atmospheric Sciences Research Center (ASRC) at the State University of New York at Albany (SUNY-Albany) for his invaluable input and suggestions, which significantly contributed to the conceptualization of the paper and improvement of its quality. Furthermore, the author acknowledges the financial assistance and computational resources generously provided by ASRC, SUNY-Albany.

\bibliographystyle{unsrt}  
\bibliography{references}

\end{document}